\documentclass[letterpaper, 10 pt, conference]{ieeeconf}
\IEEEoverridecommandlockouts
\usepackage{cite}
\usepackage{url}
\usepackage{amsmath,amssymb,amsfonts}
\usepackage{algorithmic}
\usepackage[dvipdfmx]{graphicx} % PDFの表示のため
\usepackage{textcomp}
\usepackage{xcolor}
\def\BibTeX{{\rm B\kern-.05em{\sc i\kern-.025em b}\kern-.08em
    T\kern-.1667em\lower.7ex\hbox{E}\kern-.125emX}}
    
\newif\iffigure
\figuretrue

\newcommand{\argmin}{\mathop{\rm arg~min}\limits}
\begin{document}

\title{
Bimanual Shelf Picking Planner Based on Collapse Prediction
}

\author{Tomohiro Motoda$^{1}$, Damien Petit$^{1}$, Weiwei Wan$^{1,2}$, and Kensuke Harada$^{1,2}$
	\thanks{$^{1}$Graduate School of Engineering Science, Osaka University, Japan
		%{\tt\small albert.author@papercept.net}%
	}
	\thanks{$^{2}$National Institute of Advanced Industrial Science and Technology (AIST), Japan
				%{\tt\small b.d.researcher@ieee.org}%
	}
}

\maketitle

\begin{abstract}
	In logistics warehouse, since many objects are randomly stacked on shelves, it becomes difficult for a robot to safely extract one of the objects without other objects falling from the shelf. 
	In previous works, a robot needed to extract the target object after rearranging the neighboring objects.
	In contrast, humans extract an object from a shelf while supporting other neighboring objects. 
	In this paper, we propose a bimanual manipulation planner based on collapse prediction trained with data generated from a physics simulator, which can safely extract a single object while supporting the other object.
	We confirmed that the proposed method achieves more than 80\% success rate for safe extraction by real-world experiments using a dual-arm manipulator.
\end{abstract}

%%%%% START  MAIN CONTENT
\section{Introduction}
\label{sec:introduction}
% ---
	In logistics warehouses, we often have to extract a single object that is wedged between other objects on a shelf, which is potentially dangerous for heavy objects to fall and injure human workers. 
	In this case, when a robot tries to extract one of the objects, it has to consider the positional relationship of overlapping objects and manipulate them accordingly. 
% ---
	So far, various approaches have been proposed to extract an object from a shelf. In \cite{Nam2020, Li2016, Lee2019} different methods are proposed but require a series of rearrangement operations. 
	In other cases, extraction and support relations are analyzed between pairs of objects from 3D visual perception\cite{Mojtahedzadeh2015}. 
	However, in all previous approaches, a robot extracts the target object after rearranging its neighboring objects. 
	
	% --- Fig.1 Top
	\iffigure
	\begin{figure}[tb]
		\begin{center}
			\centering\includegraphics[width=0.95\linewidth]{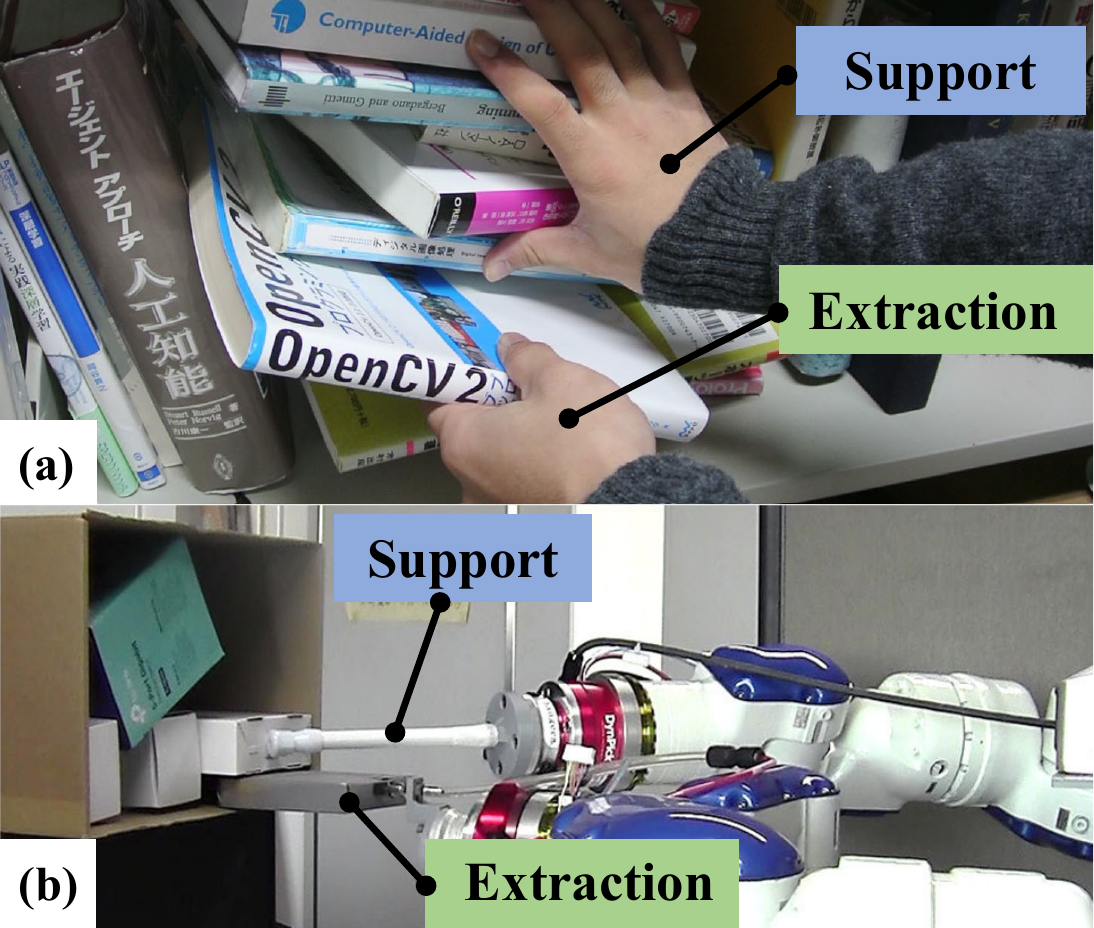}
			\caption{Extracting the target object while supporting others: (a) a human is extracting a book from the shelf while supporting the neighboring books, (b) robotic bimanual manipulation for safely extracting an object from the shelf.  } %%% MODIFIED
			\label{fig:top}
		\end{center}
	\end{figure}
	\fi
	
	% ---Fig.2 Overview of proposed method
    \iffigure
	\begin{figure*}[tb]
		\begin{center}
			\includegraphics[width=0.90\linewidth]{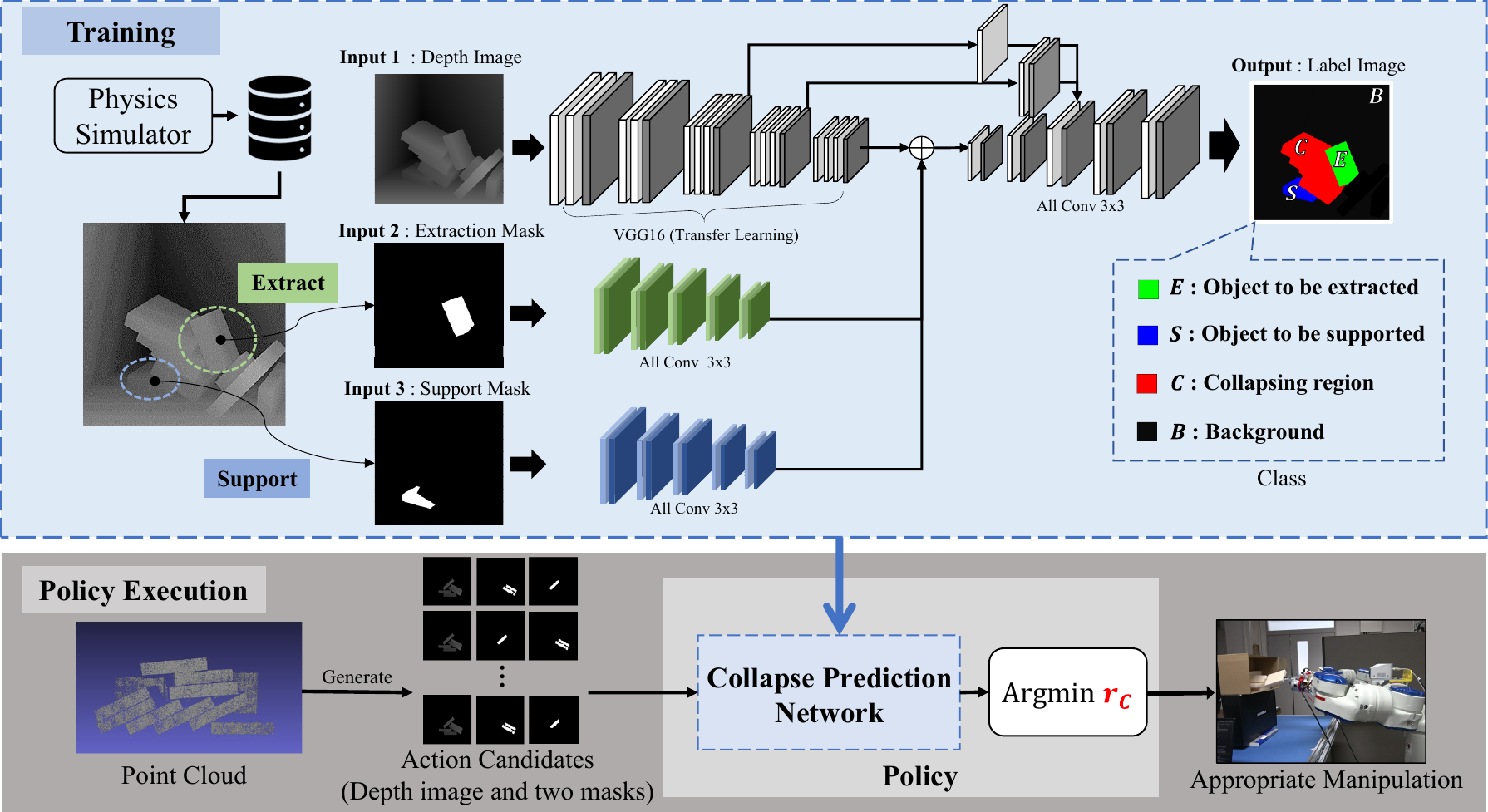}
			\caption{Schematic overview of the proposed manipulation policy based on Fully Convolutional Network (FCN). The inputs are a depth image and two binary masks. The model encodes the depth image with the VGG-16 network and two masks with a five-layer network, and concatenates these three networks. We focus on the collapsing region, $C$ (highlighted in red), in the output of the network. The method uses the argmin of $r_C$ (ratio of $C$ in the image), to return the appropriate action. The size of the depth image and its related two masks are $256\times 256$. }
			\label{fig:overview}
		\end{center}
	\end{figure*}
	\fi	

% ~*~*~ 
	Humans however, extract an object from a shelf while supporting other neighboring objects as shown in Fig.~\ref{fig:top} (a).  
	Based on this observation, we propose a bimanual manipulation planner to extract a target object from a shelf while supporting the other object as shown in Fig.~\ref{fig:top} (b). 
	% --
	To extract an object from a pile without collapse, we need to determine which of the target's neighboring object the robot have to support.
% ~*~*~  
	We propose a learning-based approach on extracting the target object from the pile while supporting the objects.
	% ---
	A network model based on a Fully Convolutional Network (FCN)\cite{fcn} has been designed to predict the pile state while extracting the target object with a pixel-wise collapse probability map.
	% ---
	The inputs of the network are a depth image of the shelf content, and two binary masks corresponding to 
	the two objects selected for extraction and support. The output of the network model is a labeled image predicting the collapsing region while the target object is extracted. If the output includes large collapsing region, we judge that such selection of supported object is not better.
	% ---
	Given this output, the robot can select the proper object to support by defining the ratio of the predicted collapsing region as the safety index to the shelf picking. 
	% ---
	In addition, to generate a large number of training data of depth images, related binary masks and label images, we use a physics simulation of the piled objects and of the extraction/support action.  
	% ---
	We experimentally verify the effectiveness of our proposed method by using a real dual-arm manipulator. We show that the robot can safely extract the target object from a shelf with a success rate larger than 80\%. By using our proposed method, we do not need to rearrange the objects placed on a shelf to extract the target object and so we increase the picking efficiency.
	
	% ---- That needs revising
	Our main contributions are:
	\begin{itemize}
    \item A Fully Convolutional Networks to infer the pixel-wise probability map of the collapsing region while extracting a selected object from a shelf (Subsection~\ref{sec:prediction_network}).
    \item A physics simulation that generate the necessary training data for the FCN (Subsection~\ref{sec:data_generation}).
    \item A robotic system able to extract a target object from a pile, in a shelf, without rearranging its surrounding objects
    \end{itemize}
    
% ---- Topic of each section
	This paper is organized as follows. In Section~\ref{sec:related_work}, we discuss the related work on shelf picking. In Section~\ref{sec:proposed_method} our proposed shelf picking method is described including a detailed explanation of the network model and its implementation. In Section~\ref{sec:experiments} 
	we describe the experimental setup. In Section~\ref{sec:discussion}, we discuss the result of our approach and experiment. Finally, the paper closes with conclusions and future work.

\section{Related Work}
\label{sec:related_work}
% ---
    In this section, we introduce some related works on object picking in 
    logistics warehouse. This topic has been extensively researched.
    There have been some works done on picking an object stored in a box, such as \cite{dexnet, Deng2019}.
%  --- Shelf Picking
    Among them, we focus on studies extracting an object from a randomly piled objects on a shelf, such as \cite{Nam2020, Li2016, Lee2019, Mojtahedzadeh2015}.
	Temtsin et al.\cite{Temtsin2017} ranked each object using a measure based on the geometrical relationships of objects and extracted an object with high rank. Mojtahedzadeh et al.\cite{Mojtahedzadeh2015} and Wu et al.\cite{Wu2020} proposed methods to learn the motion of robots in stacked or scattered environments. 
	Some researches achieved manipulation in clutter based on partially observable Markov decision processes (POMDPs). For example. Pajarinen et al.\cite{Pajarinen2014} used iterative picking/observation to disassemble the cluttered objects based on the POMDPs. To pick an object from a shelf, Li et al.\cite{Li2016} used POMDP to find the target object by rearranging the objects on the shelf safely and efficiently.
	Zhang et al.~\cite{Zhang2018, Zhang2019} used a Convolutional Neural Networks (CNNs) to estimate the order of extracting the overlapped objects by using the graph representation of the objects' position. Grotz et al.~\cite{Grotz2019} determined the order of objects to be manipulated by taking into account their support relations. 
	However, all these methods used to extract a target object from a shelf need to repeatedly rearrange the overlapping objects of the target before the extraction to avoid a collapse. 
	As far as the authors know, there has been no research on bimanual manipulation planning to extract the target object while supporting its neighboring objects at the same time, in spite of its efficiency.  

\section{Shelf Picking Method Implementation}
\label{sec:proposed_method}
    % -*-*-*-
	In this paper, we propose a bimanual manipulation method to extract a target object from a pile while supporting the other object. %\textcolor{red}{overlapping objects}. 
	In order to first verify the effectiveness of our new approach we assume that the robot achieves the task by pulling a box-shaped object out horizontally. Assuming a situation in which the insertion of fingers between objects is difficult for the robot, one arm is mounted with a suction gripper to extract the target object. The other arm has a rod-shaped end-effector to support other objects as seen in Fig.~\ref{fig:top}. We use a depth sensor to provide a 3D point cloud captured from a front point of view of the shelf containing the pile of object.

	Fig.~\ref{fig:overview} illustrates the flow of our overall architecture. The user selects the object subject to extraction, then a FCN is used to predict which objects will be affected during the extraction (I.e. collapsing region). 
	
	In the following subsections, the different steps are explained in detail. 
	
    %     % --
    % 	First, we use a physics simulator to generate different depth images, their corresponding masks and labelled images as ground truth data.
    % 	% --
    %     Second, we train our proposed prediction network using the dataset generated by the simulator.  
    % 	%The input of the network are a depth image, a mask containing the object to be extracted and a mask containing the object to be supported. The output is a image annotated with a collapsing region which indicates which objects will be affected after the extraction. 
    % 	%
    % 	%Finally, the robot uses the predicted output as a safety index to select appropriate extraction and support for manipulation. In the following subsections, the different steps are explained in detail. 
    % 	% --
    % 	Finally, the predicted output is used to compute a safety index among the different objects that can be used as a support to extract the target object. The support object which minimize the risk of collapse is then chosen (Thanks to the safety index) and the bimanual manipulation begins. One arm constrain the support object while the other arm extract the target object. In the following subsections, the different steps are explained in detail. 

	% -- DATA
	\subsection{Physics Simulator for Data Generation}
	\label{sec:data_generation}
		In this subsection, we describe the setup of the physics simulation system used for data generation.

		\subsubsection{Scene Generation}
			\iffigure
    		\begin{figure}[tb]
    			\begin{center}
    				\includegraphics[width=0.90\linewidth]{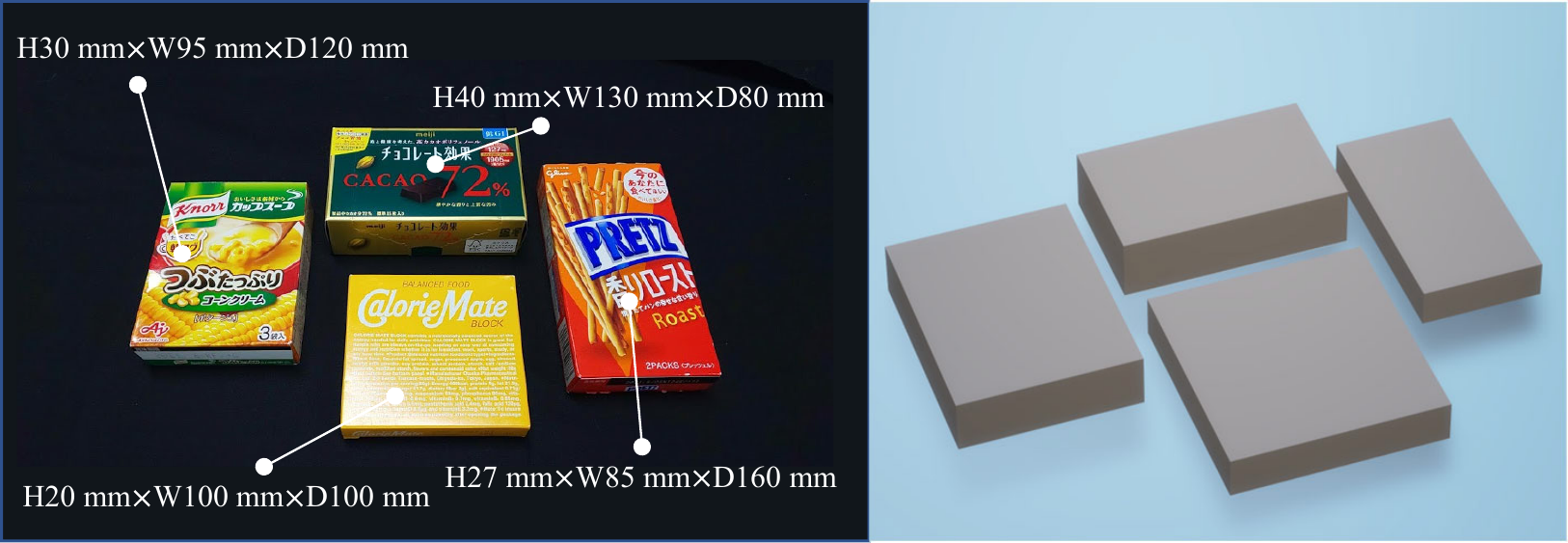}
    				\caption{Types of objects used in the simulations. The left actual boxes. The right shows the 3D models used in the simulator.}
    				%(a) This is used for simulations only with one type of objects, (b) This is used for simulations with various-size objects.
    				\label{fig:simulator_objects}
    			\end{center}
    		\end{figure}
    		\fi
			We generate a randomly stacked state of objects in the simulator. In this study, we use PhysX\footnote{https://developer.nvidia.com/gameworks-physx-overview}, i.e., a physics simulator, to configure and simulate the environment.
			% --- Parameters
            % 			\textcolor{red}{
            %     			Our simulator is built under the following setting: 
            %     			We set the coefficient of static and dynamic friction to be 0.9 and 0.8, respectively. We also set the coefficient of restitution to be 0.1, and the density to be 1.0 $kg/m^3$. We perform the simulation by placing six box-shaped objects randomly on a shelf where the size shelf is 300mm$\times$300mm$\times$500mm. 
            % 			}
			% +++++++MOTODA CORRECTION (3/17)+++++++++++++++
		    Our simulator is designed with the following settings. 
		    Considering a situation where many product boxes are on a shelf, thus we set the simulation parameter referred to the actual movements of them. For both objects and a shelf in our environment, we set the coefficient of static and dynamic friction to be 0.9 and 0.8, respectively. We also set the coefficient of restitution to be 0.1 and the density to be 1.0 $kg/m^3$. 
		    We perform the shelf picking simulation by placing six objects from a set of objects. As the number of objects on a shelf increases, the extraction generally becomes more difficult. In our study, we fix the number of the objects to be six, which can generate the successful cases empirically in about 50\% even if the target object for extraction/support is randomly selected. Moreover, for the after-mentioned verification, we prepared two sets of objects: One type of object (H~20 mm$\times$ W~100 mm$\times$ D~100 mm), and four objects of various sizes (the height is 27 -- 40 mm, the width is 85  -- 130 mm, and the depth is 80 -- 160 mm), illustrated in Fig.~\ref{fig:simulator_objects} for the detail. We generate the dataset with either of the sets according to the conditions. 
		    
		    % +++++++++++++++++++++++++++++++++++++++++++++
		    % MEMO ++++++++++++++++++++++++++++++++++++++++
    			% - MEMO Material setup -
    			% * Box   - Coefficient of static friction : 0.9f
    			% * Box   - Coefficient of dynamic friction : 0.8f
    			% * Box   - Coefficient of restitution : 0.1f
    			% * Box   - density : 1.0f [kg/m^3]
    			% * Shelf - Coefficient of static friction : 0.9f
    			% * Shelf - Coefficient of dynamic friction : 0.8f
    		    % * Shelf - Coefficient of restitution : 0.1f
    			% Drop objects arranged at even intervals, which initial pose are random. 
    			% same-size object -> 6 (randomly)
    			% various-sizes object -> 6 <- randomly
    			% Box size (same-size simulation) -> 100x100x20 [mm]
    			% Box size (various-size simulation) -> 130x80x40,160x85x25, 100x100x20, 120x95x30
    			% (H~40mm$\times$ W~130mm$\times$ D~80mm, H~27mm$\times$ W~85mm$\times$ D~160mm, H~30mm$\times$ W~95mm$\times$ D~120mm)
    			% bin space : 300x300
    			% VHACD (Volumetric Hierarchical Approximate Convex Decomposition) : Collision Check
    			%The details are explained in the next section.

		\subsubsection{Data generation and Simulation Procedure}
			 \iffigure
    		\begin{figure}[tb]
    			\begin{center}
    				\includegraphics[width=0.95\linewidth]{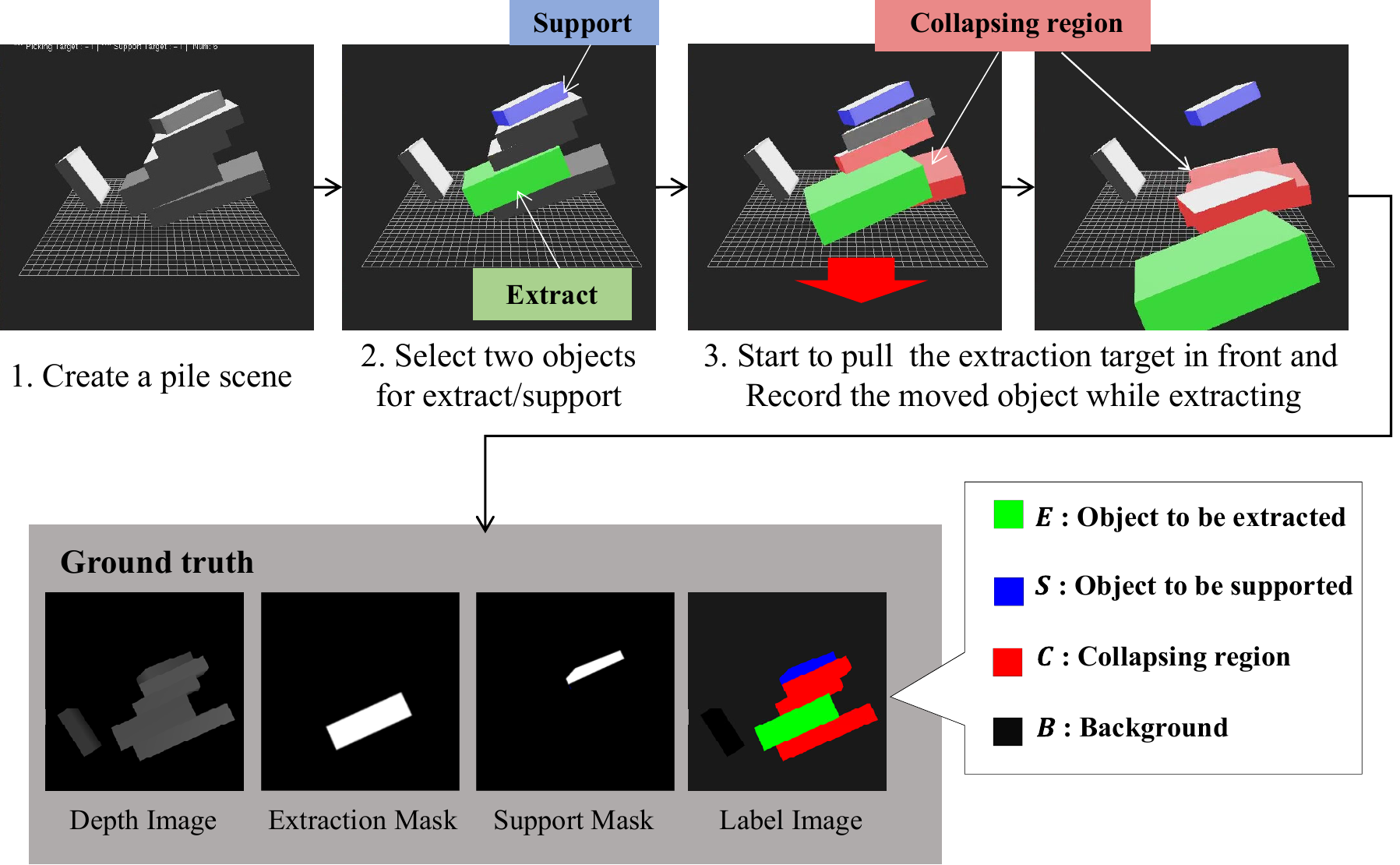}
    				\caption{
    				    Dataset generation procedure with physics simulator. The upper row shows a scene of a simulation while the lower row shows the images of ground truth generated from the simulated scene. 
    				}
    				% Dataset generated with physics simulator. (a), (b) Failed attempts to extract an object without objects collapsing, (c), (d) Successful cases for extracting an object. They can select two targets to support and extract correctly.
    				\label{fig:dataset}
    			\end{center}
    		\end{figure}
    		\fi		
    		
		    Fig.~\ref{fig:dataset} shows the process of simulation.
		    First, the simulator creates a pile scene.
			Second, one object moves horizontally in the simulation. Here, we assume that the robot pulls out one object from the shelf horizontally toward the observer. The other object is supported, and we assume that it remains stationary, hence, it is not affected by interference or gravity, and its pose does not change. 
			Finally, in case there is some change in any object pose other than the two targets, we record these objects as collapsed subject to entangling/collision. 
			In one simulation, we can obtain the tuple, consisting of three images as input data and one labeled image as output data, as shown in Fig.~\ref{fig:dataset}. 
			
			% We do not consider the end-effector during the task execution so that we avoid unbalanced outputs subject to external factors, such as interference of end-effectors. 
			%Furthermore, the objects' collapse is judged only by the objects in the vicinity of the target. A large collapse misleads the network learning for manipulation, and it is challenging to select the appropriate objects to extract and support correctly. 

    % -- Network
    \subsection{Collision Prediction Network}
    \label{sec:prediction_network}
		This subsection describes the neural network that predicts the objects affected by the target object extraction and so most likely to fall or collapse.
    	Similar to the fully convolutional network (FCN) used in~\cite{fcn}, our model classifies each pixel belonging to the collapsing region. 
		
		\subsubsection{Ground Truth}
    	    The input data consist of a depth image ($256\times 256$) and two binary masks ($256\times 256$). One mask is an object for extraction, and the other mask is an object for support.
    	    % ---
    		The output data is a prediction of the classifications for each pixel of the image ($256\times 256$). We define four classes: Object to be extracted $E$, Object to be supported $S$, Collapsing region $S$, and Background region $B$, as shown in Fig.~\ref{fig:dataset}. 
    %       We define four classes as follows:  
    % 		\begin{itemize}
    % 			\item $E$: Object to be extracted.
    % 			\item $S$: Object to be supported.
    % 			\item $C$: Collapsing region. %Objects that move and rotate after extraction.
    % 			\item $B$: Background region. 
    % 		\end{itemize}
    		The collapsing region $C$ expresses the region of objects which move or fall from the shelf while extracting the target object. These input/output data are automatically generated from the simulator. 
		
		\subsubsection{Network Architecture}
		    Our network model consists of an encoder for extracting the feature value of the input and a decoder for producing the segmented image at its original resolution. Fig.~\ref{fig:overview} illustrates the network architecture. 
    	    First, the encoder part consists of three networks. The model generates the feature maps for a depth image with the VGG-16 network\cite{vgg} pre-trained by ImageNet\cite{imagenet_cvpr09} and for two masks, each with five convolution layer network. These three outputs are then concatenated into one feature map.
        	Next, the decoder part with five convolution layers up-sample the feature maps to the original resolution with deconvolution. Moreover, our network uses the skip architecture by referring to prior examples\cite{fcn, unet2015}, which combines the feature maps of the lower layers with those of the upper layers to recover the general location information while preserving the local information. 
		
	% Manipulation
	%\subsection{Manipulation Strategy}
	\subsection{Manipulation Planning}
		This section describes the manipulation procedure to perform the extraction by applying the trained network. 
		The robot acquires point clouds from a depth sensor installed in front of the shelf and generates three input images from this observation. One is a depth image converted from the point clouds to the depth map. The other two images are mask images representing the object to be extracted and the object to be supported. The mask image, $M_{c}$, is a binary image from each cluster, $c_i \ (i=0, 1, 2, ..., N-1)$ of point clouds, which is classified by object segmentation based on the region growing method\cite{region_growing} and the binarization. Fig.~\ref{fig:segmentation} shows the process.       In the actual experiments, the robot end-effectors approach each object toward the center of gravity in these masks. 
		
		%% --- MOTODA (3/17) ---
        		We can define action candidates $A$ according to use situations: 
        		(1) the other situation is that the robot chooses the safest pair of extraction/support objects (for example, empty a shelf). In this case, we define action candidates $A$ by preparing all the combinations of two different targets of the extraction/support action.
        		(2) one is that we need to extract a predetermined target object. In this case, in advance the user selects an object subject to extraction arbitrarily, we define action candidates $A$ by choosing another subject to support, 
        		
        		Next, we define the safety index to select the best action from all the candidates. Here, the output shows the region, $R_E$, $R_S$, $R_C$, and $R_B$ that indicates the regions of four different classes ($E$, $S$, $C$, and $B$). If $R_C$ is large, it will increase the risk of collapse. Based on this assumption, we can define the following risk index:
		    % -----------------
		    %Next, we evaluate the safety of each candidate based on the output from our proposed network. The output shows the region, $R_E$, $R_S$, $R_C$, and $R_B$ that indicates the regions of four different classes ($E$, $S$, $C$, and $B$).  
		    %We assume that if $R_C$ is large, it will increase the risk of collapse. Based on this assumption, for all images in the input dataset that are candidates for the operation, we use $R_C$ corresponding to $C$ in the output result to define the possibility of collapsing after the operation occupancy rate of $R_C$ as follow. 
		    % ------------------
		\begin{equation}
			r_c \ (a)= \frac{area(R_C^{a})}{area(R_E^{a} \cup R_S^{a} \cup R_C^{a} \cup R_B^{a})}
			\label{eq:safety}
		\end{equation}
		$R_E^{a}$, $R_S^{a}$, $R_C^{a}$, and $R_B^{a}$ denote the regions in the output of an action candidate $a$ for two selected objects. $area(\cdot)$ indicates the area of the region.
		Our algorithm selects the input data that is the smallest for index $r_c$ based on Eq.~(\ref{eq:safety}) and determines the best action, $a$, of all the action candidates to be manipulated by the robot. 
		\begin{equation}
			a = \argmin_{a' \in A}{r_c \ (a')}
		\end{equation}
		If the robot's motion is out of the control range, we eliminate it from the candidates and select the next best move. 
		
		\iffigure
		\begin{figure}[tb]
			\begin{center}
				\centering\includegraphics[width=0.95\linewidth]{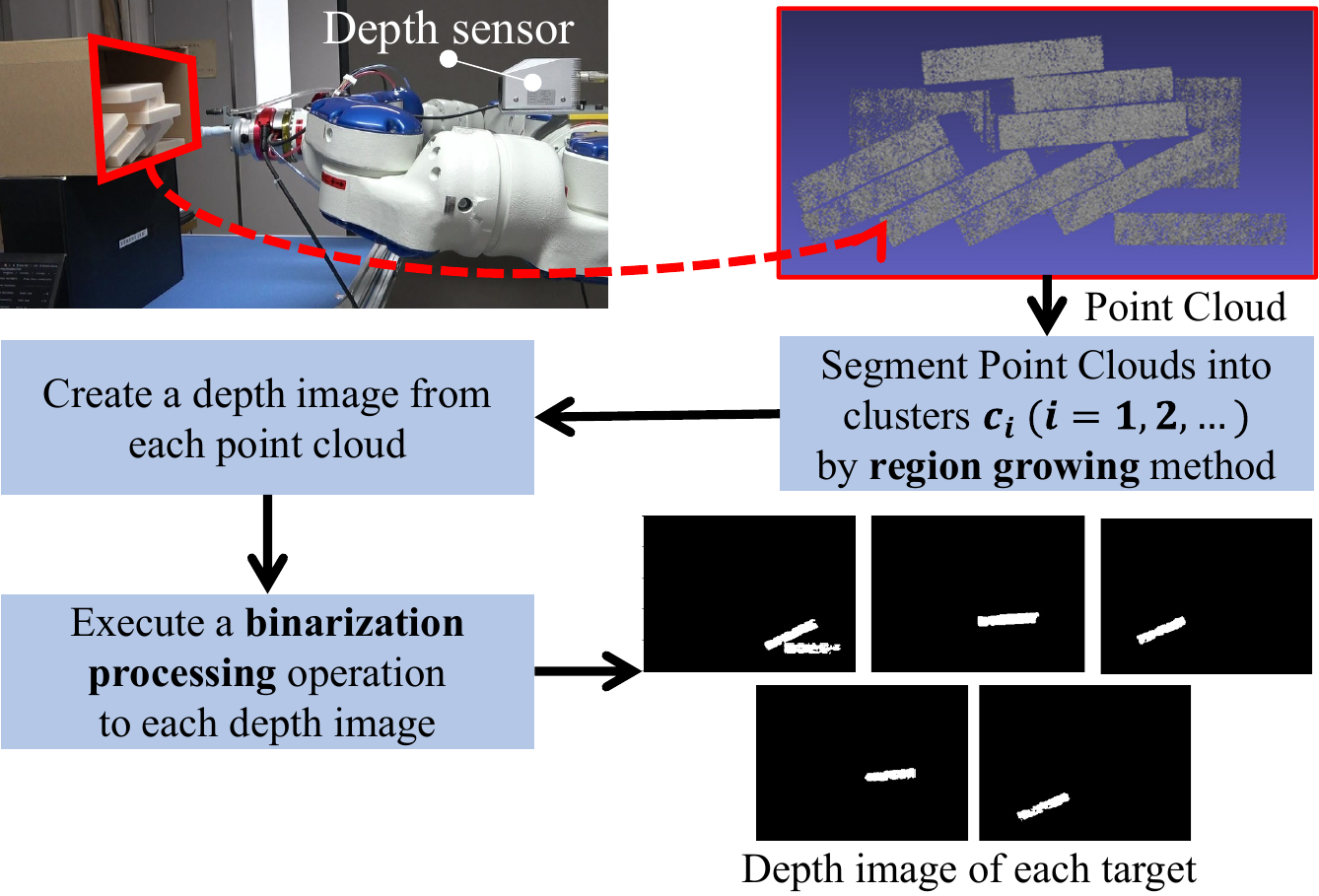}
				\caption{Process of generating candidates with the segmentation method. }
				\label{fig:segmentation}
			\end{center}
		\end{figure}
		\fi
	
\section{Experiments}
\label{sec:experiments}
	\subsection{Training Settings}
		In our experiments, we acquired 15,000 pairs of input and output images from the simulator. From these datasets, we used 90\% as training data and 10\% as validation data. We augmented our training dataset through left-right inversion and utilized the network using 30,000 pairs. We set the initial learning rate to 0.0001 up to 30 epochs and 0.00001 from 30 epochs onward. The batch size was 1, and we use the Adam\cite{Adam} as the optimizer. The number of epochs during training was 50, and each epoch required 27,000 iterations. 
		In our training, we used the NVIDIA RTX 2060 super (8 GB VRAM). 
		
	\subsection{Experimental Setup}
		\iffigure
		\begin{figure}[tb]
			\begin{center}
				\includegraphics[width=0.90\linewidth]{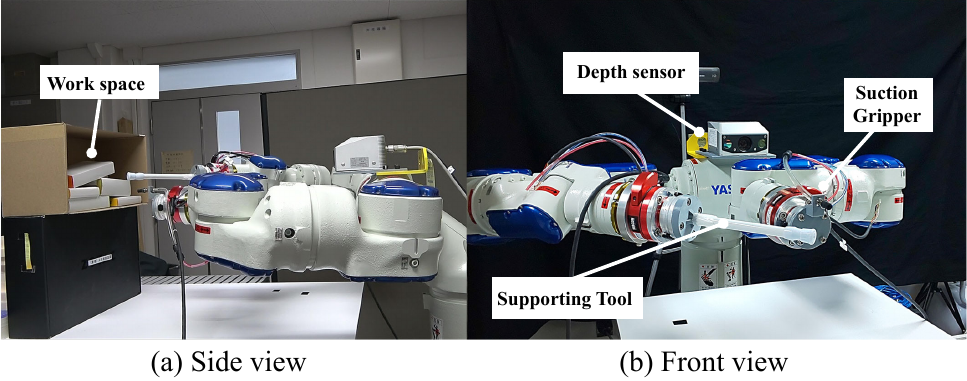}
				\caption{Experimental setup.} %  with the robot MOTOMAN-SDA5F (right-hand: rod-shaped end-effector, left-hand: suction gripper), and depth camera YCAM3D-III-10L. 
				\label{fig:settings}
			\end{center}
		\end{figure}
		\fi
		To verify the effectiveness of the proposed method, we conducted experiments using an actual robot under several conditions. Fig.~\ref{fig:settings} shows the experimental environment used in the verification. 
		We use the MOTOMAN-SDA5F (Yaskawa Electric Corp.)\footnote{https://www.motoman.com/en-us/products/robots/industrial/assembly-handling/sda-series/sda5f}, a bimanual robot with 7 degrees-of-freedom robot arms, which has a suction gripper and a plastic rod-shaped end effector (the bar's length is 20 cm) at the tip of each arm of the robot. 
		The YCAM3D-10L (YOODS Co.,~Ltd.)\footnote{https://www.yoods.co.jp/products/ycam.html}, a 3D depth sensor, is installed at the front of the shelf. 

	%\subsection{Experimental Conditions}
	\subsection{Results}
		\iffigure
		\begin{figure}[tb]
			\begin{center}
				\includegraphics[width=0.85\linewidth]{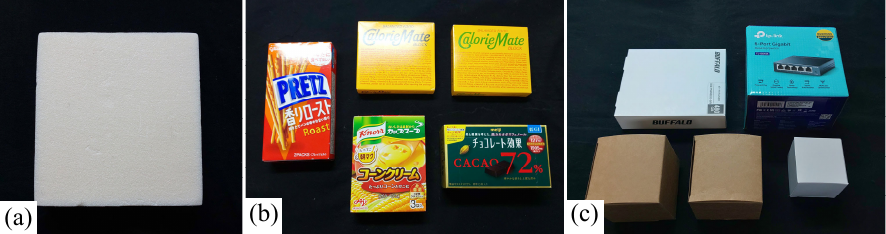}
				\caption{Objects used to evaluate the generalization of the proposed method: (a) experiments for objects of the same size, and experiments for objects of various sizes, and (c) experiments for new objects (not used in simulations).}
				\label{fig:objects}
			\end{center}
		\end{figure}
		\fi
		\iffigure
		\begin{figure}[tb]
			\begin{center}
				\includegraphics[width=0.90\linewidth]{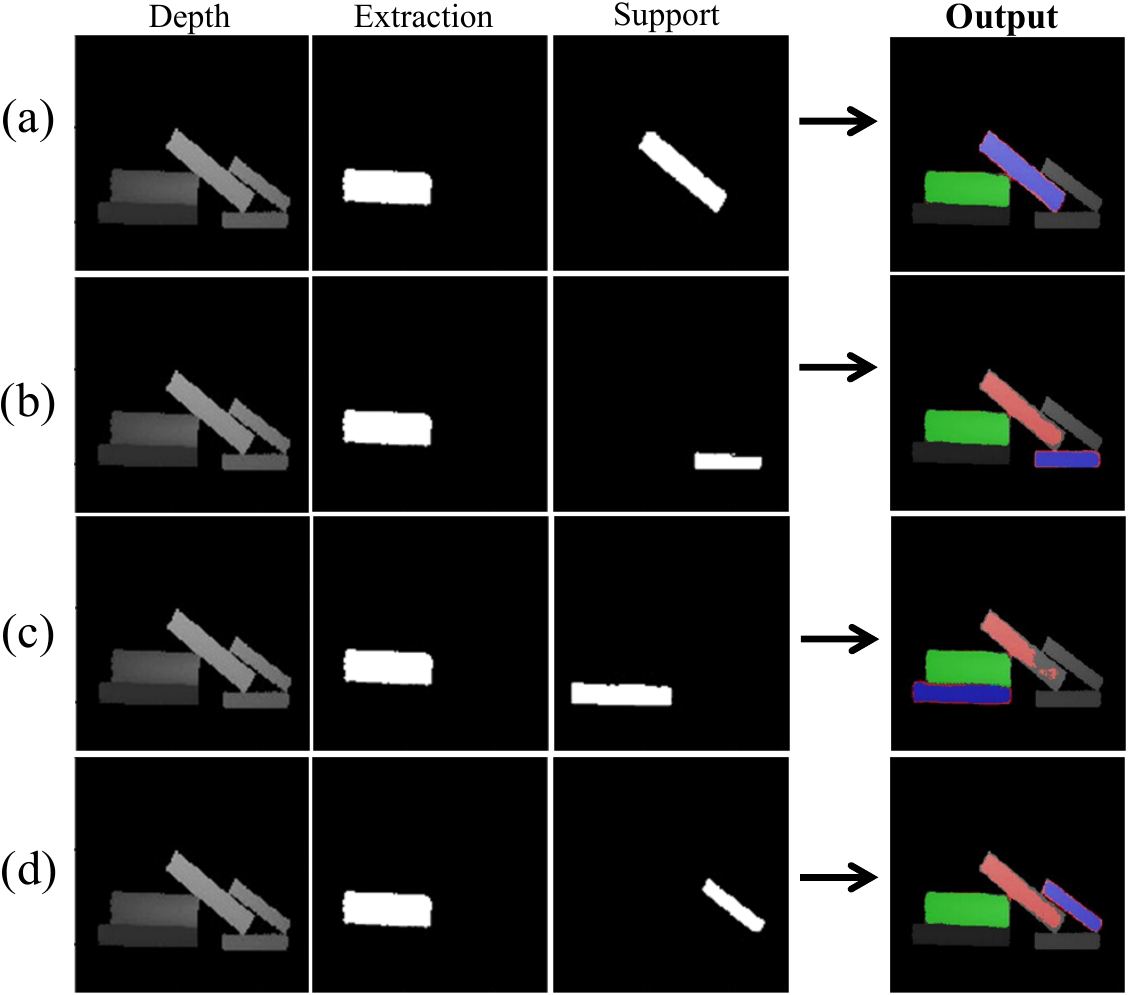}
				\caption{
				    Visualization results from the proposed network model: (a)-(d) Outputs of extraction for a target object while supporting different objects. The red area on the images shows the predicted collapse region ($R_C$). The green region shows the object to be extracted correspond to $R_E$, and the blue region shows the object to be supported correspond to $R_S$. 
				}
				\label{fig:outputs}
			\end{center}
		\end{figure}
		\fi
	    \iffigure
		\begin{figure}[tb]
			\begin{center}
				\includegraphics[width=0.9\linewidth]{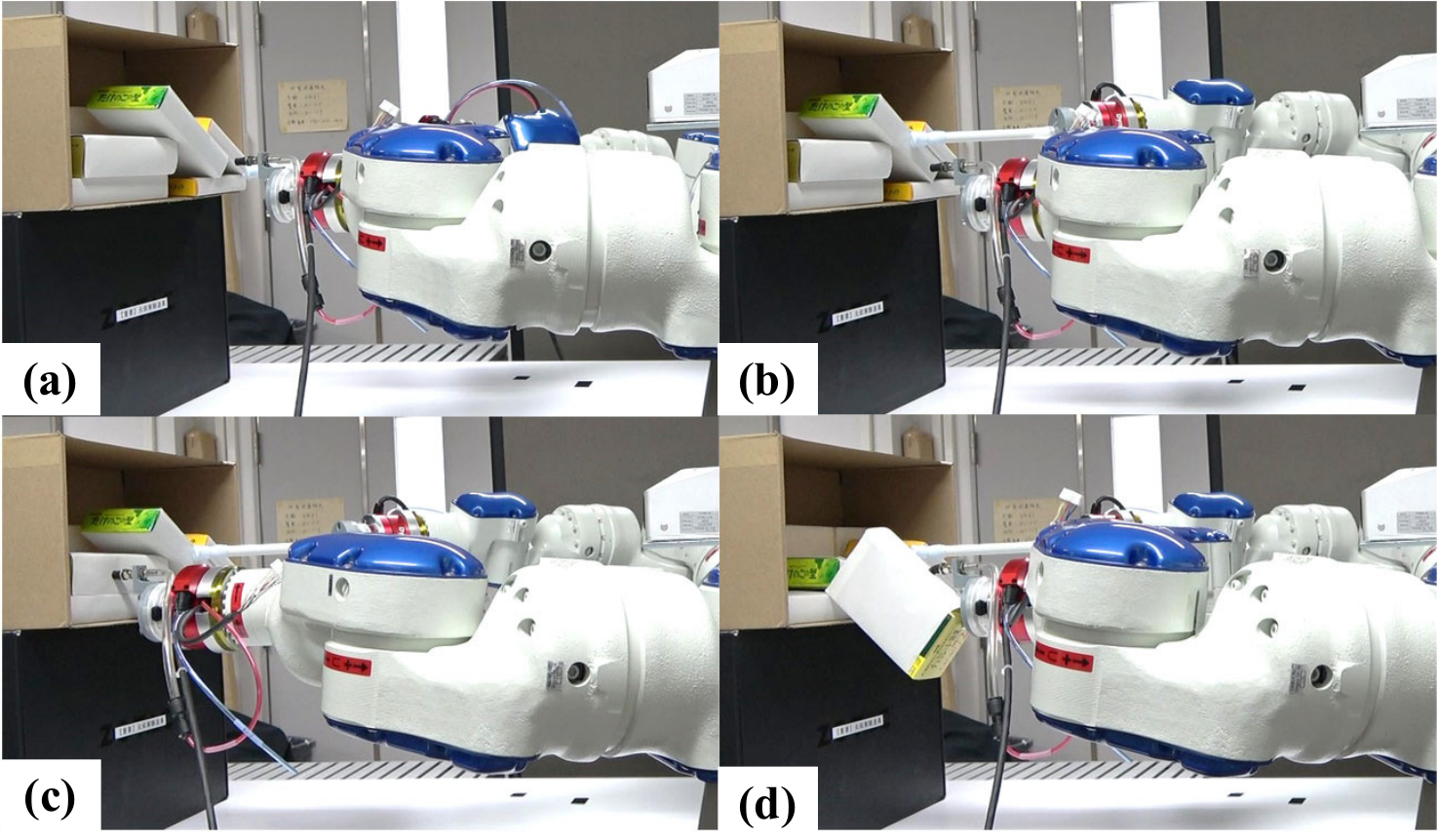}
				\caption{
    				Description of target manipulation with the proposed method: (a)-(d) a series of scenes in one task. 
				}
				\label{fig:sequence}
			\end{center}
		\end{figure}
		\fi
		\iffigure
		\begin{table}[tb]
			\caption{Experimental result}
			\begin{center}
				\includegraphics[width=0.9\linewidth]{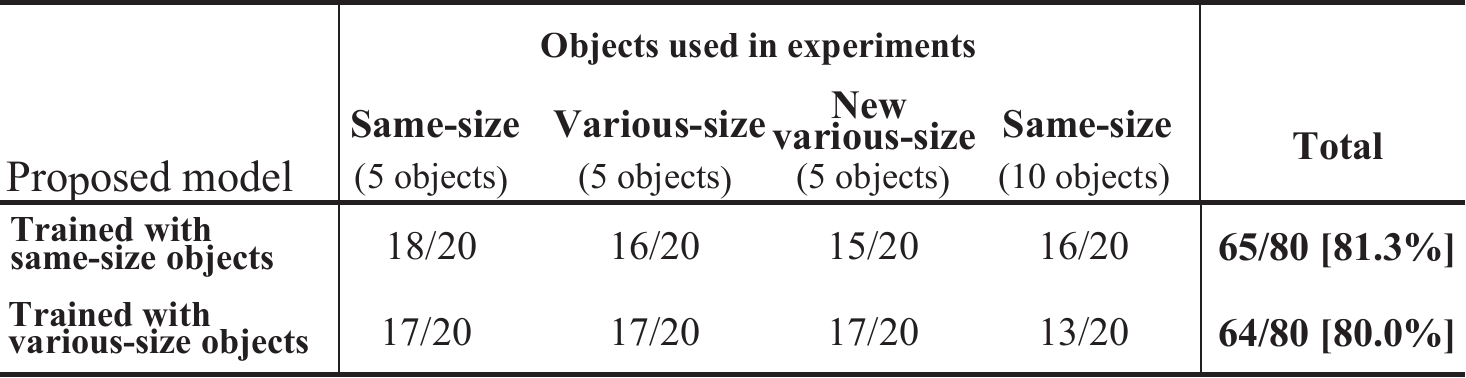}
				\label{tab:ex_result}
			\end{center}
		\end{table}
		\fi
	
	    % MOTODA 3/17 --- 
        To evaluate the performance by using the actual robot, we consider experiments on two use situations.
        
        \subsubsection{Choosing the safest pair of extraction/support object} % (assuming task to empty a shelf)
        \label{subsubsec:safest}
          We verify the performance of our prediction network through experiments that the obot always chooses the safest action.
         % -----
    		In our real-world experiments, we used the target objects as shown in Fig.~\ref{fig:objects}(a), (b), which is the same size as models used in our simulations (Fig.~\ref{fig:simulator_objects}). Moreover, in order to evaluate the generalization capability, we separately prepared new objects (Fig.~\ref{fig:objects}(c)).
    		We used the following conditions in experiments: 
    		\begin{itemize}
    		    \item 5 objects of same size, as shown in Fig.~\ref{fig:objects}(a).
    		    \item 5 objects of various sizes, as shown in Fig.~\ref{fig:objects}(b).
    		    \item 5 new objects of various sizes (not used on the simulator), as shown in Fig.~\ref{fig:objects}(c).
    		    \item 10 objects of same size, as shown in Fig.~\ref{fig:objects}(a).
    		\end{itemize}
            Furthermore, we prepared two network models trained with different datasets generated from the simulations with objects of the same size or various sizes.   
    		With each trained model, we conducted 20 trials in every four patterns by changing the size or number of objects. If the robot removes only one object from the shelf, we regard it as a success; otherwise, we consider it failed. 

    		As shown in Table \ref{tab:ex_result}, we conducted robotic experiments under the above mentioned conditions. The robot achieved the high success rate across all conditions, and the overall extraction success rate were 81.3\% (65/80) for the model trained with objects of the same size  and 80\% (64/80) for the model trained with the dataset of various sizes. 
    		
    	\subsubsection{Extracting a predetermined target object} % (assuming use in logistics warehouse) 
        \label{subsubsec:extract}
          We assume that a specific target object is needed in a practical situation. 
          The user chooses one object to extract from a shelf in advance, and in that case, our policy determines which is the object to support correctly.
          
          We show the output of our network as Fig.~\ref{fig:outputs}(a)-(d) in the case. The robot selects the best action from the output that region $R_C$ (highlighted in red) is small as Fig.~\ref{fig:outputs} (a). Fig.~\ref{fig:sequence} shows the experimental scene. When the correct action is selected, the robot first presses down the one object with the stick on the right-hand and then pulls out the target with the left-hand suction gripper. Based on the learning results, we confirmed that the robot selected combinations of objects that are less likely to collapse and execute the safest manipulation. 
          
          Moreover, we conduct 20 trials in that case. At each trial, the object to be extracted is not changed. It should be noted that we trained the network with the dataset generated in the simulations using various-size objects (Fig.~\ref{fig:simulator_objects}) in this verification. In 20 trials of experiments under this conditions, the robot can extract single target object without collapse in success rate of 85\% (17/20). We confirm that our network works well. 
		
	%\subsection{Results}

\section {Evaluation}
\label{sec:discussion}
     In this section, the performance is evaluated concerning two points. (1) We set a benchmark of the prediction performance based on segmentation metrics and compare our proposed network under different conditions, (2) we acquire the success rate, representing the percentage of the completed when extracting a single target object without collapse by using a real robot.  

    \subsection{Prediction Performance}
    \label{subsec:cnn_evaluation}
    
    	% --- about Metrics 
    	We confirm the network can correctly predict the collapsing regions with ground-truth data, as shown Table~\ref{tab:model_evaluation}.
		To evaluate the performance of the collapse prediction, we focus only on the collapsing region $C$ in this study. Our metrics include $precision$, $recall$, and $IoU$ calculated in pixels, between the predicted and ground-truth. We calculate the average values on metrics with a hundred ground-truth data, and compare two networks trained with different training datasets. 
		Moreover, to verify a generalization of the performance, we prepare the ground-truth in two different patterns: target objects of same size or various sizes . We empirically set the threshold of classification for each pixel to 0.4. 
	
		\iffigure
		\begin{table}[tb]
			\caption{Trained Model Evaluation.}
			\begin{center}
				\includegraphics[width=0.9\linewidth]{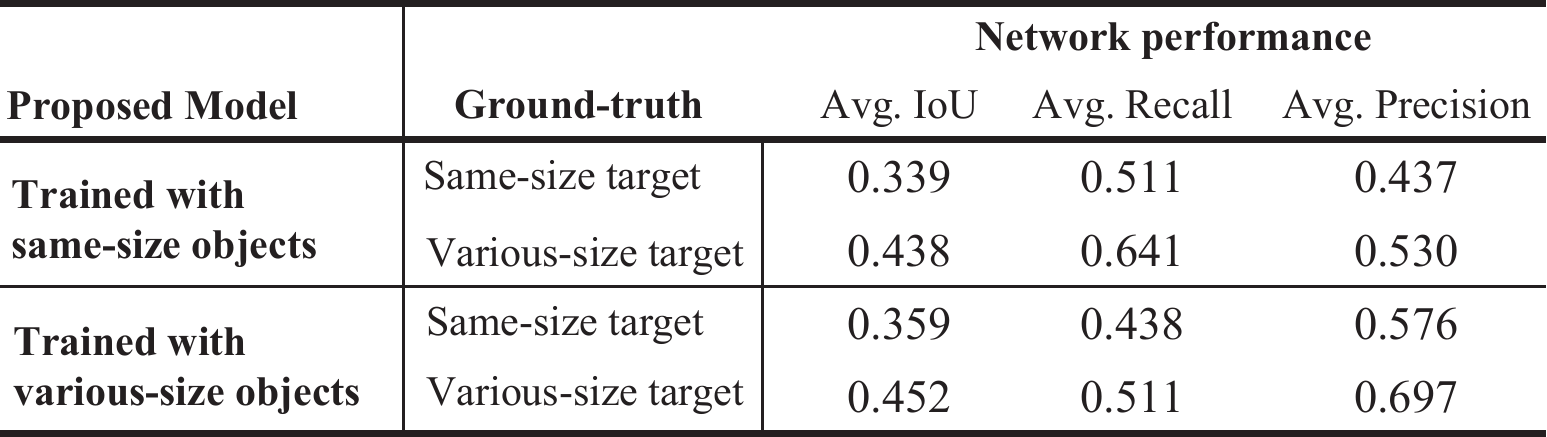}
				\label{tab:model_evaluation}
			\end{center}
		\end{table}
		\fi
		
		% --- about Analysis
		As shown in Table~\ref{tab:model_evaluation}, even when we use networks with different training datasets, there is no significant difference on each metric to the same ground-truth. This result indicates that the size of the object has little effect on learning. 
		In contract, when we use the network trained with the objects of various sizes , $IoU$ and $precision$ increase in both ground-truth data. By using our method, the collapsing region tends to become a shape similar to the object model. It is assumed that the network trained with objects of same size is relatively sensitive to shape differences. 
		Therefore, training with objects of various sizes works well for correctly predicting the region. 
		
	\subsection{Real-world Manipulation}
	    
	    As shown in Table~\ref{tab:ex_result}, the robot extracted successfully up to 81.3\% (65/80) for the same object dataset and 80\% (64/80) for the dataset of objects of different sizes. The success rate of each object is not significantly affected in different datasets. 
	    Similarly, there is no difference in the success rate when the objects are the same (Fig.~\ref{fig:objects}(a)) and when the size of the objects is randomized (Fig.~\ref{fig:objects}(b)). The success rates of 75\% (15/20) and 85\% (17/20) were confirmed in the experiments with objects of new various sizes (Fig.~\ref{fig:objects}(c)), indicating that there was no overfitting of our learning results.
	
	    In extracting a predetermined target object, our method achieved a high success rate of 85\%, indicating that our method can work well in logistics warehouse. Moreover, the success rate is almost equal to other experimental results. We showed that it was possible to make good predictions even when the target object to be extracted was limited.

		% -- bad case and bad result
		In some failed cases, the robot executed intuitively incorrect actions, such as supporting an object unrelated to extracting a target. This indicates a problem with simulator settings. For example, there are trials where the robot can remove one target safely without supporting the other object. We also consider that the robot misunderstood some uncertain manipulations as successful trial. It is necessary to reconstruct the dataset or evaluate each action on each successful trial.  
		% ---
		As shown in Table~\ref{tab:ex_result}, the success rate decreased in ten objects of the same size. For example, if an object is not simply put on another object, the robot needs to support more than two objects. In our method, however, the robot can only support one object,  causing a low success rate. In our future work, we will address this issue.

    \subsection{Discussion}
    % -*-*-*-
		We proposed the learning-based approach to predict collapse after extracting one object while supporting the other one. 
		The conventional learning-based approaches\cite{Zhang2018, Zhang2019} predict the support relationship as Section~\ref{sec:related_work} mentioned. However, considering the complex pile, it will become more difficult to determine the support object by geometry shape and/or physical interaction. 
		In contrast, our proposed method can directly predict whether the selected action is proper or not without checking the complex structure. 
		However, in order to realize the new idea, we focused only on box-shaped objects for the sake of simplicity. Our future work will be extended to more complex-shaped objects and apply them to daily life. 

\section{Conclusions}
\label{sec:conclusions}
% --- Summary
    This paper described a shelf picking method for safely extracting a single object from a shelf while supporting the other object. By using our proposed network model that predicts the objects that would collapse, a bi-manual robot was able to extract the object without objects falling. 

% --- 展望
    In the future, we plan to make improvements on support actions and our simulator. 
    In particular, we will analyze the trial result on each simulation by adding actions to support and extract in an appropriate way for object types.

%%%%%% FINISH MAIN CONTENT

%\section*{Acknowledgement}
%We would like to thank Editage (www.editage.com) for English language editing. 

%\bibliographystyle{./bibliography/IEEEtran}
%\bibliography{./bibliography/IEEEabrv,./bibliography/IEEEexample}

\vspace{12pt}
\end{document}